\documentclass{midl} 

\usepackage{mwe} 
\usepackage{svg}
\usepackage{booktabs}
\usepackage{comment}
\jmlrvolume{-- Under Review}
\jmlryear{2025}
\jmlrworkshop{Full Paper -- MIDL 2025 submission}
\editors{Under Review for MIDL 2025}

\title[Towards Fair Medical AI: Adversarial Debiasing of 3D CT Foundation Embeddings]{Towards Fair Medical AI: Adversarial Debiasing of 3D CT Foundation Embeddings}

\midlauthor{\Name{Guangyao Zheng\nametag{$^{1,2}$}} \Email{gzheng7@jhu.edu} \\
\Name{Michael A. Jacobs\nametag{$^{1,3,4,5}$}}\Email{Michael.A.Jacobs@uth.tmc.edu} \\
\Name{Vladimir Braverman\nametag{$^{1,2,6}$}} \Email{vb21@rice.edu} \\
\Name{Vishwa S. Parekh\nametag{$^{3,7}$}} \Email{vishwa.s.parekh@uth.tmc.edu} \\
\addr $^{1}$ Department of Computer Science, Rice University, Houston, TX, USA \\
\addr $^{2}$ Department of Computer Science, The Johns Hopkins University, Baltimore, MD, USA \\
\addr $^{3}$ Department Of Diagnostic And Interventional Imaging, McGovern Medical School, UTHealth Houston, Houston, TX, USA, \\
\addr $^{4}$The Russell H. Morgan Department of Radiology and Radiological Science, The Johns Hopkins University School of Medicine, Baltimore, MD 21205\\
\addr $^{5}$ Sidney Kimmel Comprehensive Cancer Center, The Johns Hopkins University School of Medicine, Baltimore, MD 21205\\
\addr $^{6}$ Google Research, Mountain View, CA, USA\\
\addr $^{7}$ Department of Neurosurgery, The Johns Hopkins University School of Medicine, Baltimore, MD, USA  \\
}
\begin{document}
\maketitle
\begin{abstract}
Self-supervised learning has revolutionized medical imaging by enabling efficient and generalizable feature extraction from large-scale unlabeled datasets. Recently, self-supervised foundation models have been extended to three-dimensional (3D) computed tomography (CT) data, generating compact, information-rich embeddings with 1408 features that achieve state-of-the-art performance on downstream tasks such as intracranial hemorrhage detection and lung cancer risk forecasting. However, these embeddings have been shown to encode demographic information, such as age, sex, and race, which poses a significant risk to the fairness of clinical applications. 

In this work, we propose a Variation Autoencoder (VAE) based adversarial debiasing framework to transform these embeddings into a new latent space where demographic information is no longer encoded, while maintaining the performance of critical downstream tasks. We validated our approach on the NLST lung cancer screening dataset, demonstrating that the debiased embeddings effectively eliminate multiple encoded demographic information and improve fairness without compromising predictive accuracy for lung cancer risk at 1-year and 2-year intervals. Additionally, our approach ensures the embeddings are robust against adversarial bias attacks. These results highlight the potential of adversarial debiasing techniques to ensure fairness and equity in clinical applications of self-supervised 3D CT embeddings, paving the way for their broader adoption in unbiased medical decision-making. The code is available at \url{https://github.com/BioIntelligence-Lab/VAE-Adversarial-Debiasing}.
\end{abstract}

\begin{keywords}
Fairness, Imaging Databases, Medical Image Compression, Progressive Streaming, Data-Efficiency
\end{keywords}

\section{Introduction}

The rapid advancements in self-supervised learning have revolutionized medical imaging, enabling the extraction of compact, information-rich embeddings from large-scale unlabeled datasets. These embeddings have demonstrated exceptional performance across various clinical applications, including detecting intracranial hemorrhages and forecasting lung cancer risk \cite{yang2024advancing,CT_Foundation_Demo}. However, recent studies have highlighted a critical limitation: these embeddings often encode sensitive demographic information, such as age, sex, and race \cite{zheng2024demographic}. This unintended encoding introduces risks of bias in downstream clinical tasks, potentially compromising fairness and equity in medical decision-making \cite{gichoya2022ai}. Such biases can lead to disparities in patient outcomes, perpetuating systemic inequities in healthcare. Addressing these challenges is imperative to ensure that AI models used in medical imaging support equitable healthcare delivery.

\begin{figure}
    \centering
    \includegraphics[width=\textwidth]{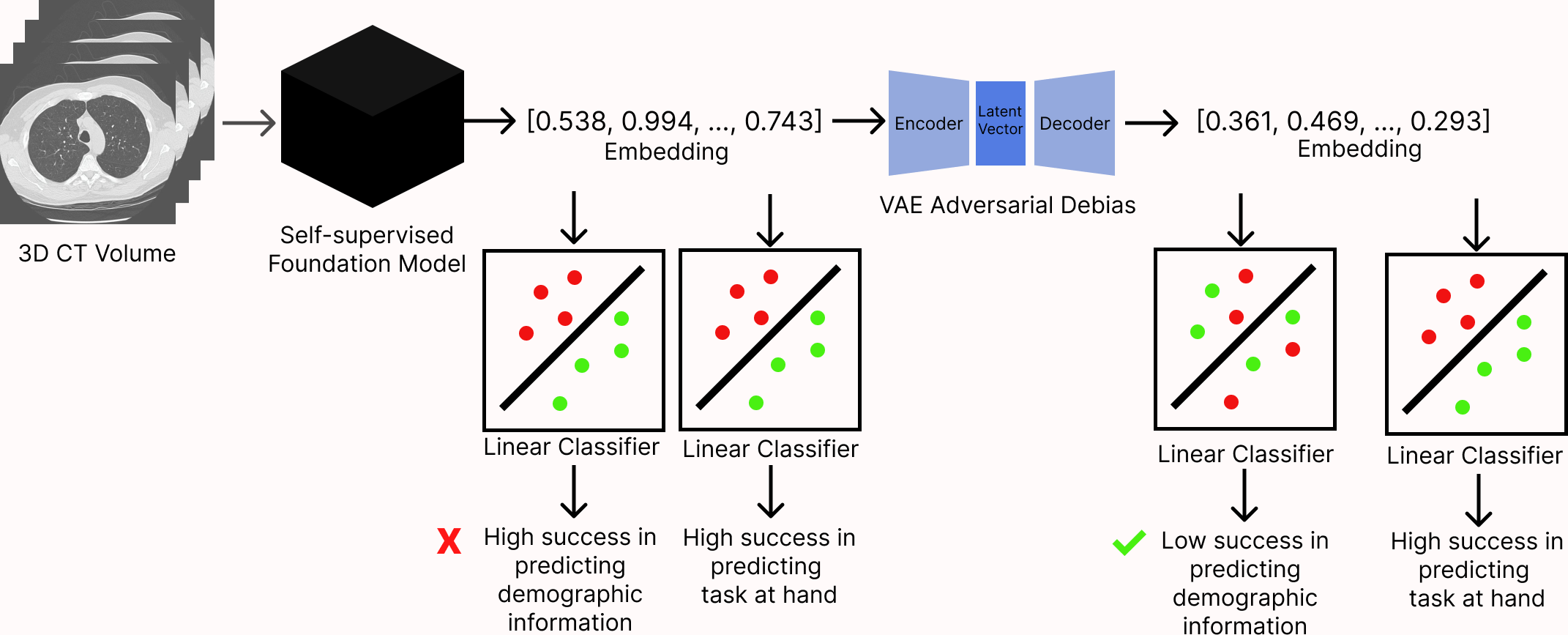}
    \caption{Conceptual illustration of our framework. The original 3D CT Foundation Model embedding encodes demographic information, which can lead to bias in downstream tasks. Our VAE can debias multiple demographic information while preserving downstream task performance.}
    \label{fig:concept}
\end{figure}

To mitigate these risks, we propose an adversarial debiasing framework designed to transform 3D CT embeddings into a new latent space that excludes sensitive demographic information while preserving their utility for downstream clinical tasks. Adversarial debiasing is a technique used in machine learning to mitigate bias in models, especially when dealing with sensitive attributes like gender, race, or age. It leverages adversarial learning to encourage the model to make predictions that are not influenced by these sensitive attributes while maintaining overall performance on the main task \cite{zhang2018mitigating}. In this work, we extend this approach using a variational autoencoder (VAE)-based architecture to transform demographic-encoding embeddings to demographic-free embeddings without affecting the performance of downstream models, as shown in Figure \ref{fig:concept}.

Adversarial debiasing has been widely studied in the literature as a key technique for mitigating bias in machine learning models, particularly in medical applications \cite{correa2024efficient,correa2021two,jin2024debiased,agarwal2024debias,yang2023adversarial}. However, existing approaches often fail to meet three key constraints: independence from downstream tasks, the ability to debias multiple sensitive attributes simultaneously, and compatibility with black-box foundation models. For instance, Debias-CLR \cite{agarwal2024debias} employs contrastive learning but requires separate models for each sensitive attribute, while DNE \cite{jin2024debiased} masks sensitive information in images using adversarial noise without explicitly evaluating its impact on downstream performance. Other frameworks, such as those by Yang et al. \cite{yang2023adversarial} and Correa et al. \cite{correa2021two, correa2024efficient}, integrate adversarial debiasing directly into task-specific models, limiting generalizability and necessitating access to internal model representations. In contrast, our proposed VAE Adversarial Debiasing framework ensures downstream task independence, debiases multiple attributes simultaneously, and operates effectively on embeddings from black-box self-supervised models, making it a more adaptable and clinically practical solution.

We evaluate the proposed framework on the NLST lung cancer screening dataset, focusing on two critical metrics: fairness and predictive accuracy \cite{national2011national}. First, we assess whether the transformed embeddings successfully eliminate demographic signals, ensuring that downstream models cannot infer sensitive attributes such as age or sex. Second, we measure the impact of debiasing on the predictive accuracy of lung cancer risk forecasting at 1-year and 2-year intervals.

\section{Materials and Methods}

\subsection{Dataset and 3D CT Foundation Model Embeddings}

This retrospective study utilized the publicly available National Lung Screening Trial (NLST) dataset, which contains 3D CT images of the lungs from patients aged 55–74 years, along with associated demographic information such as age, sex, and race \cite{national2011national}. The dataset was specifically curated for lung cancer screening and represents a diverse patient population. Table \ref{tab:data} provides an overview of the demographic distribution within the dataset.

To extract embeddings, we employed the CT Foundation model, a state-of-the-art self-supervised learning framework designed to encode 3D CT images into compact, information-rich feature representations \cite{yang2024advancing, Kiraly_Traverse_2024, CT_Foundation_Demo}. Using this tool, we obtained embeddings for all CT scans in the NLST dataset. The embeddings consist of 1408-dimensional feature vectors, capturing critical spatial and contextual information from each image. For model development and evaluation, we adhered to the established patient-wise data splits provided by the CT Foundation tool: training data included 10,299 patients (52,696 images), while test data consisted of 2,199 patients (11,421 images). These splits ensured no patient overlap between training and test sets, maintaining the integrity of the evaluation.

\subsection{Demographic Prediction}

To evaluate whether the embeddings produced by the CT Foundation model encode demographic information, we conducted a series of experiments using linear classifiers, which are simple yet effective for detecting the presence of demographic information without the risk of overfitting or introducing complex transformations. For the classification of sex, we utilized a sigmoid classifier, which provides a direct measure of the classifier’s ability to predict sex from the embeddings. For predicting age, we used linear regression, given the continuous nature of age as a demographic feature. Both models were trained on the embeddings from the training set, and their performance was evaluated on the test set. To assess the performance of the classifiers, we used accuracy and the AUC (Area Under the Curve) of the ROC (Receiver Operating Characteristic) curve, which provides a more comprehensive measure of classifier performance. The AUC score offers insight into the model's discriminative ability. Statistical significance for evaluating these models was set at $p < 0.05$.

\subsection{Adversarial Debiasing}

To remove demographic information from the 3D CT embeddings while maintaining their effectiveness for clinical prediction tasks, we employed an adversarial debiasing framework based on a Variational Autoencoder (VAE) architecture. 
The VAE framework is particularly effective for learning continuous, disentangled representations, which is crucial for both debiasing and preserving task-relevant information.

The VAE architecture consists of two main components: the encoder and the decoder. The encoder uses a linear layer to map the input 3D CT embeddings into a latent space of mean and log variance parameters. These parameters are reparameterized to allow backpropagation during training. The decoder comprises a linear layer that reconstructs the embeddings from the latent space. During training, the VAE loss is computed as the sum of two components: (1) the reconstruction loss (mean squared error, MSELoss), which ensures the model generates embeddings that are close to the original input, and (2) the KL divergence loss, which regularizes the latent space to follow a unit Gaussian distribution. 

To introduce adversarial debiasing, we added an adversarial network to predict demographic information from the latent representations. The adversary consists of multiple branches, each corresponding to a specific demographic attribute. In the NLST dataset case, age and sex. For binary classification tasks (sex), the adversary applies a sigmoid activation to the output of a branch, followed by binary cross-entropy loss (BCELoss). For continuous variables (age), the adversary directly computes mean squared error loss (MSELoss). The total adversarial loss is the sum of the individual losses from the two branches, and the encoder is trained to minimize this loss, ensuring that the demographic attributes cannot be predicted from the embeddings.



\section{Experiments and Results}

\subsection{Experiment 1: Evaluation of baseline adversarial debiasing performance}

\textbf{Experimental Setup: } To comprehensively evaluate our debiasing framework, we compared the original embeddings with adversarially debiased embeddings across three key metrics: (1) the predictive performance of downstream tasks, specifically lung cancer prediction at 1 and 2 years; (2) the extent to which the embeddings encode demographic information, such as age and sex, assessed by training linear models to predict these attributes from both original and debiased embeddings; and (3) the level of bias in age and sex, quantified using Equal Opportunity Difference (EOD) \cite{obermeyer2019dissecting}. EOD measures disparities in model performance across demographic groups by computing the difference in true positive rates (TPRs) between groups, where a higher EOD indicates greater unfairness, and an EOD closer to zero signifies fairer model performance. By evaluating EOD on lung cancer prediction at 1 and 2 years using both the original and debiased embeddings, we assessed whether our framework successfully mitigated bias while preserving predictive accuracy. The VAE was trained for 100 epochs with a learning rate of 0.0005 for the encoder and decoder and 0.002 for the adversary, respectively, using a batch size of 32. The dimensionality of the latent space was empirically determined to be 500 (Appendix \ref{app:latent}), balancing the trade-off between representational capacity and demographic disentanglement.

\begin{figure}
    \centering
    \begin{subfigure}[]
        \centering
        \includegraphics[width=0.35\textwidth]{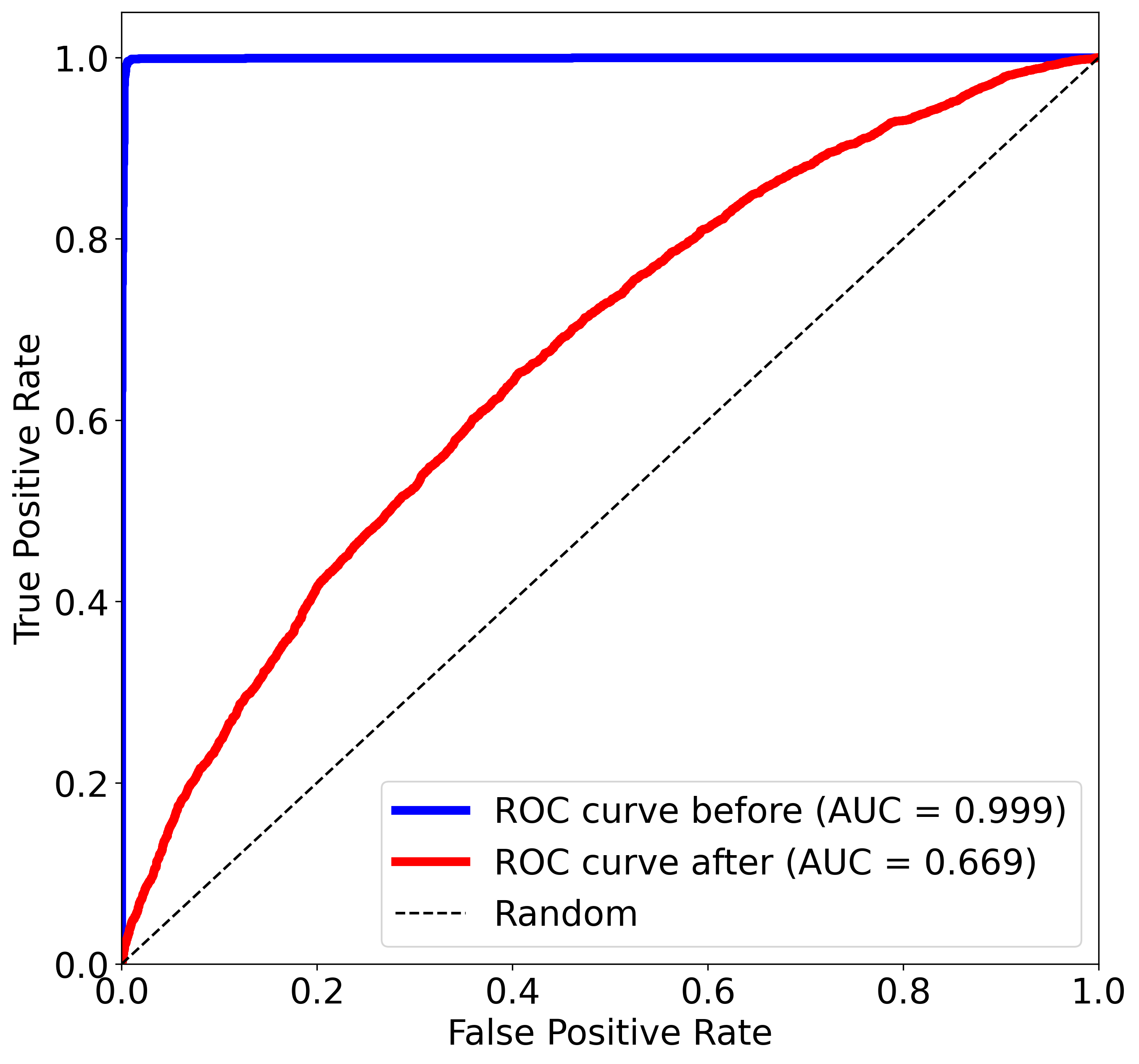}
    \end{subfigure}%
    ~
    \begin{subfigure}[]
        \centering
        \includegraphics[width=0.35\textwidth]{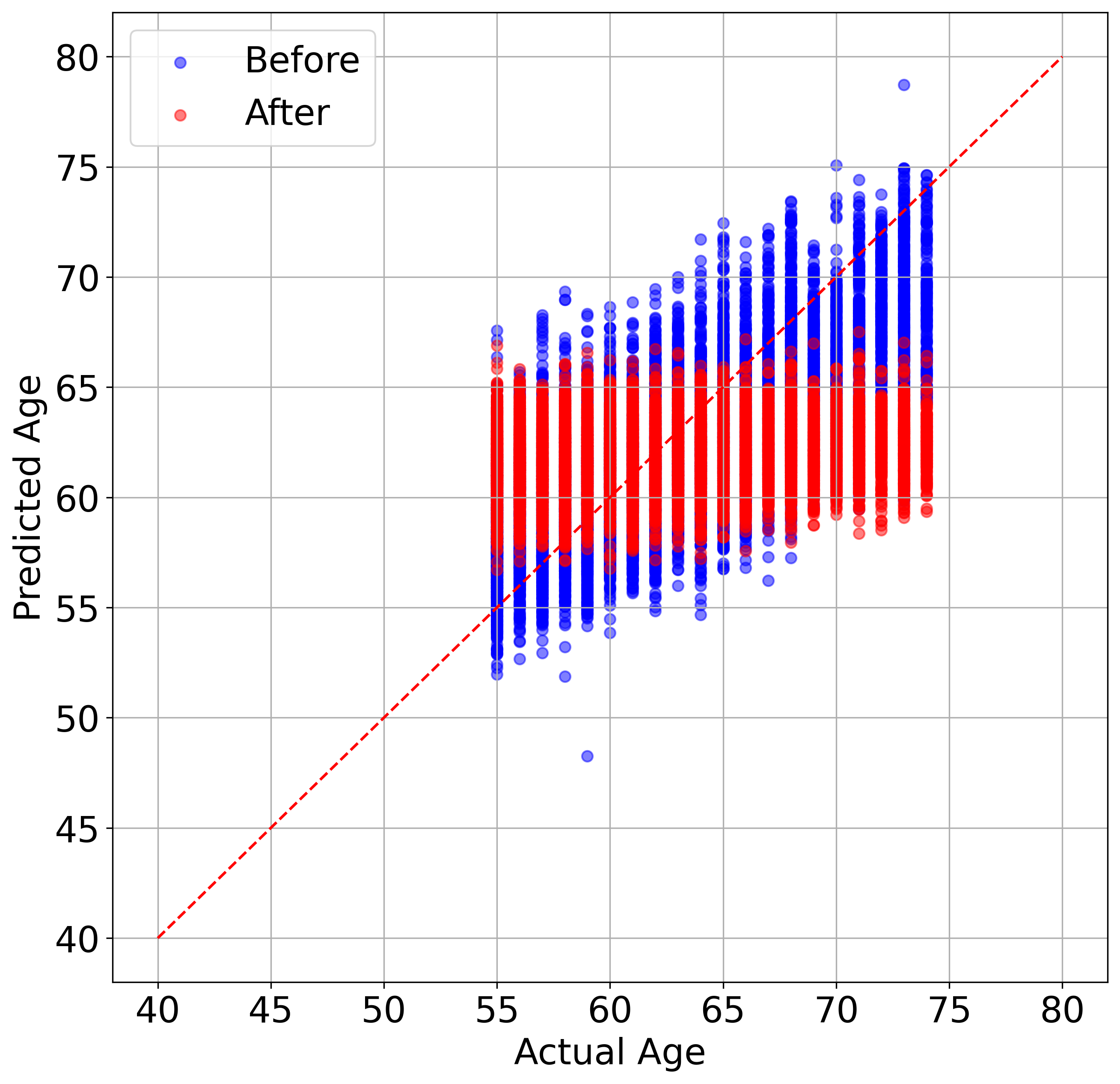}
    \end{subfigure}
    \caption{(a) ROC curve of linear classifier performance on sex prediction comparing original embedding (blue) vs. embedding after VAE reconstruction (red). (b) Scatter plot of linear classifier performance on age prediction comparing original embedding (blue) vs. embedding after VAE reconstruction (red).}
    \label{fig:demographic}
\end{figure}

\textbf{Results: } \textit{Demographic Prediction: }In the sex prediction task, the original embeddings exhibited a strong encoding of demographic information, achieving an accuracy of 0.994 and an AUC of 0.999. Following debiasing, these metrics declined substantially to 0.647 and 0.669, respectively, indicating a significant reduction in the model’s ability to infer sex from the embeddings. This trend is further corroborated by the ROC curve in Figure \ref{fig:demographic}(a), which illustrates the marked decrease in sex predictability, underscoring the efficacy of our framework in mitigating demographic bias. Similarly, for age prediction, the original embeddings yielded a mean absolute error (MAE) of 2.734, reflecting a strong correlation between age and the learned representations. After debiasing, the MAE increased to 4.169, signifying a considerable attenuation of age-related information. The scatterplot in Figure \ref{fig:demographic}(b) provides additional empirical support, demonstrating that while the original embeddings closely adhered to the red dotted line—indicating a strong linear correlation between predicted and actual age—post-debiasing, the predicted age exhibited increased variability, with values distributed within the 55 to 75 range, as represented by the rectangular region in the plot.

\begin{figure}
    \centering
    \begin{subfigure}[]
        \centering
        \includegraphics[width=0.35\textwidth]{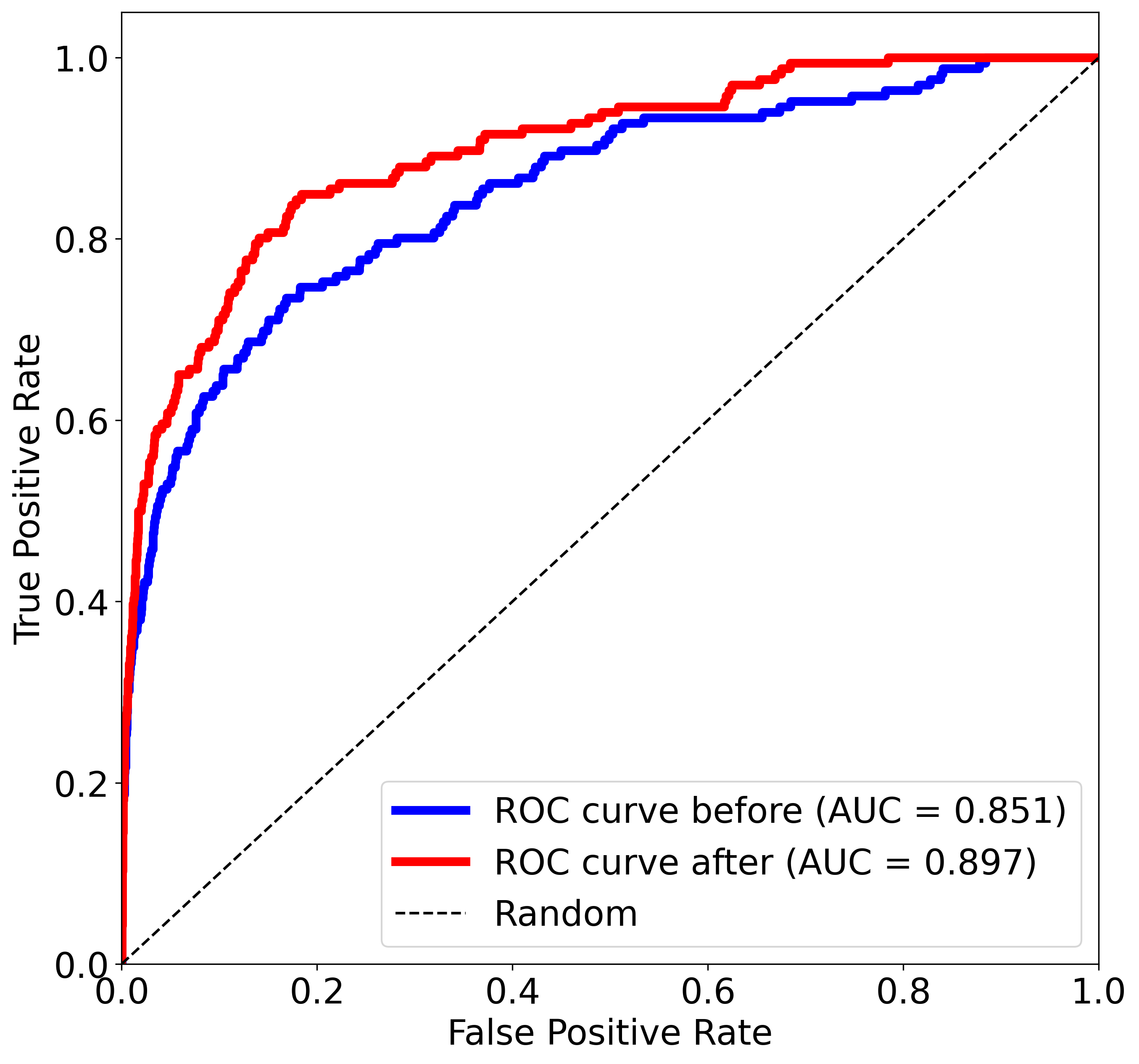}
    \end{subfigure}%
    ~
    \begin{subfigure}[]
        \centering
        \includegraphics[width=0.35\textwidth]{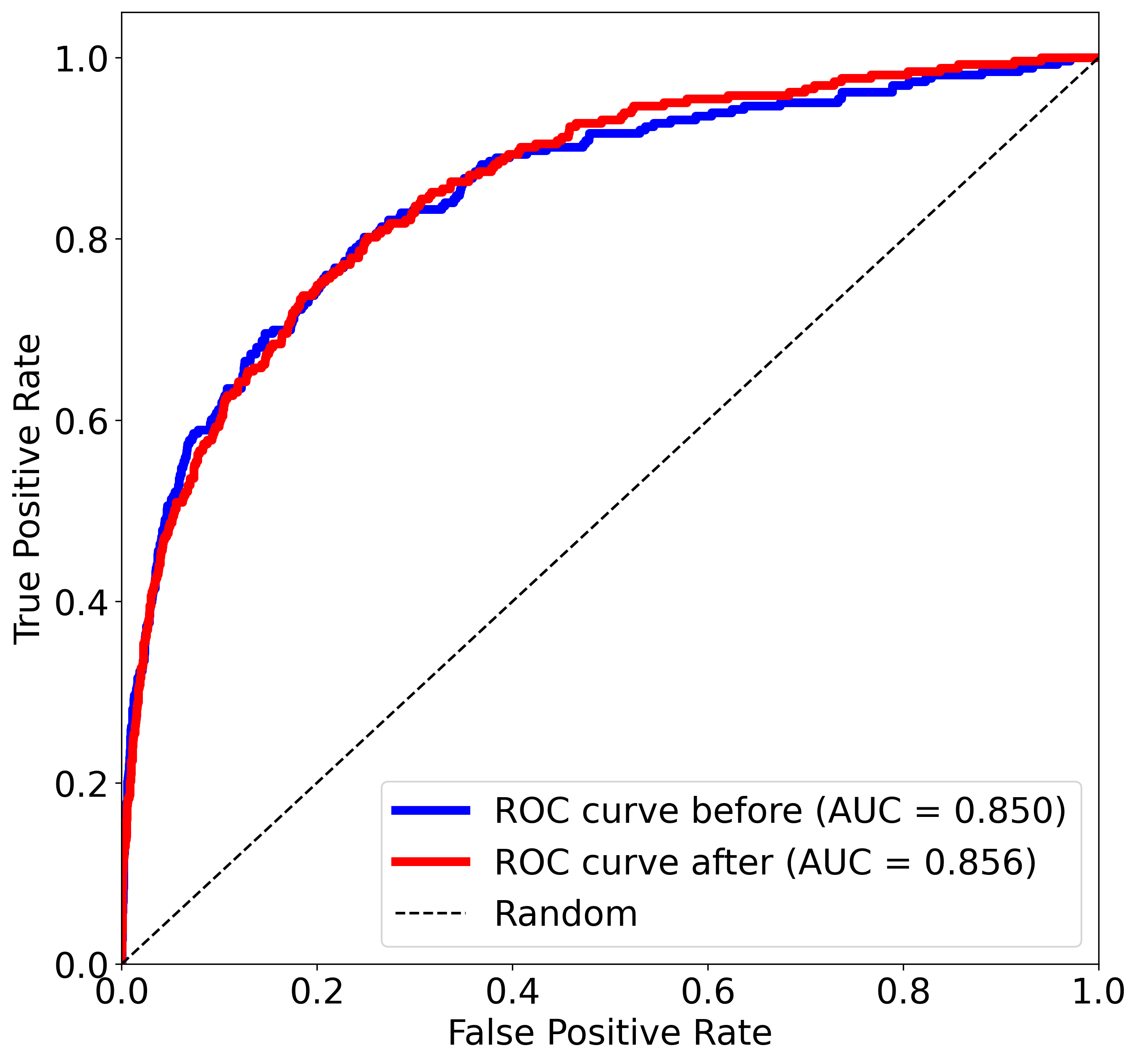}
    \end{subfigure}
    \caption{ROC curve of linear classifier performance on (a) cancer in 1 year, (b) cancer in 2 years prediction comparing original embedding (blue) vs. embedding after VAE reconstruction (red)}
    \label{fig:lungcancer}
\end{figure}

\textit{Lung Cancer Prediction: }Despite the reduction in demographic information, the embeddings retained strong predictive performance for lung cancer risk assessment. For the 1-year lung cancer prediction task, the accuracy remained unchanged at 0.986 for both the original and debiased embeddings. Similarly, for the 2-year lung cancer prediction, accuracy was nearly identical at 0.977 across both embedding types. The ROC curves in Figure \ref{fig:lungcancer} further illustrate that debiasing not only preserved performance but also led to a slight increase in the AUC scores for both 1-year and 2-year cancer prediction, reinforcing that our debiasing approach does not compromise the model’s predictive capabilities.

\textit{Downstream Task Fairness: }As shown in Table \ref{tab:EO}, our VAE-based debiasing framework effectively enhances fairness for both sex and age, as evidenced by the reduced EOD compared to the original embeddings.

These findings demonstrate that our adversarial debiasing framework effectively removes multiple sensitive demographic attributes from the embeddings while maintaining their utility for clinical prediction tasks. Furthermore, this procedure operates independently of the downstream task, ensuring that debiasing does not negatively impact lung cancer risk prediction. 

\begin{table}[htp!]
\center

\begin{tabular}{c|cc}
\toprule
Cancer in 1 year & Without debiasing & After debiasing \\
\midrule
Sex & 0.035 & \textbf{0.014} \\
Age & 0.333 & \textbf{0.111} \\
\toprule
Cancer in 2 years & Without debiasing & After debiasing \\
\midrule
Sex & 0.023 & \textbf{0.013} \\
Age & 0.200 & \textbf{0.063} \\
\bottomrule
\end{tabular}
\caption{EOD performance of without debiasing (original embedding) and after debiasing (VAE debiased embeddings) for sex and age as sensitive attributes. Evaluated on downstream tasks cancer in 1 year and cancer in 2 years. EOD lower is better.}
\label{tab:EO}
\end{table}

\subsection{Experiment 2: Robustness to Data Poisoning}

\textbf{Experimental Setup: } Demographic predictability can expose models to adversarial data poisoning, where an attacker deliberately perturbs data to degrade model performance for specific demographic groups \cite{kulkarni2024hidden}. To simulate such an attack, we selectively "target" a particular demographic group (e.g., males) by randomly flipping their lung cancer prediction labels. In this experiment, we varied the proportion of label flipping across $0\%, 25\%, 50\%, 75\%, and 100\%$ for each demographic group separately, while ensuring that the labels for the other demographic group remained undisturbed. Similar to experiment 2, we evaluated if our framework improved the robustness of downstream models to demographically targeted adversarial attacks using EOD for sex and on downstream tasks cancer in 1 year and cancer in 2 years using the original embedding and the VAE debiased embedding.

\begin{figure}
    \centering
    \begin{subfigure}[]
        \centering
        \includegraphics[width=0.44\textwidth]{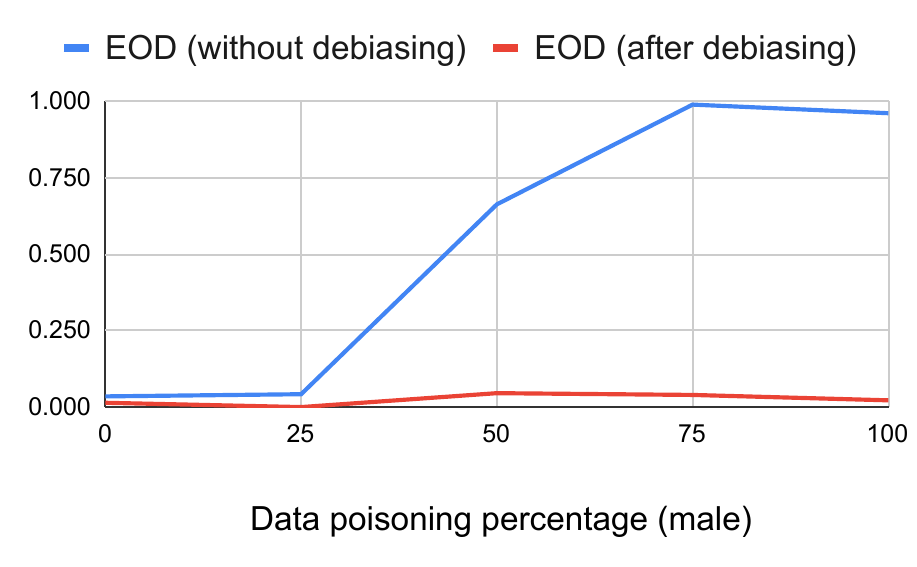} 
    \end{subfigure}%
    ~
    \begin{subfigure}[]
        \centering
        \includegraphics[width=0.44\textwidth]{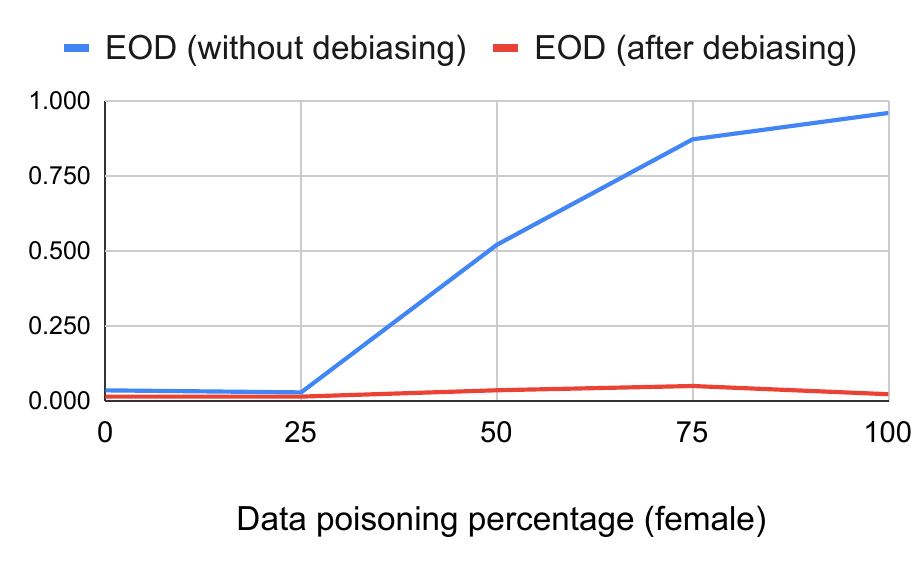}        
    \end{subfigure}
    \caption{EOD comparison for original embedding (blue) vs. VAE debiased embedding (red) for cancer in 1 year prediction when the (a) male,  (b) female patient's cancer in 1 year label is poisoned by X percentage (X = 0, 25, 50, 75, 100).}
    \label{fig:data_poison}
\end{figure}

\begin{figure}
    \centering
    \begin{subfigure}[]
        \centering
        \includegraphics[width=0.44\textwidth]{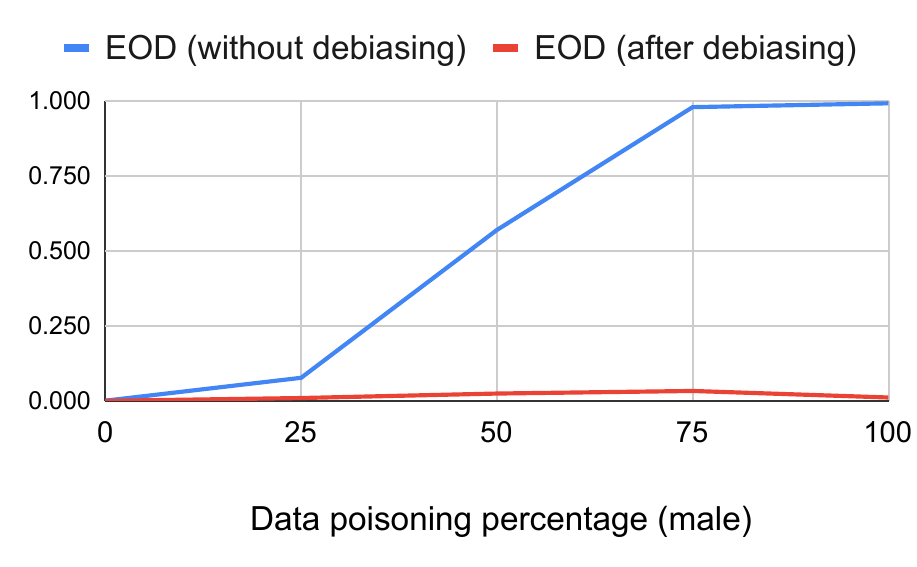}
    \end{subfigure}%
    ~
    \begin{subfigure}[]
        \centering
        \includegraphics[width=0.44\textwidth]{year1female.pdf} 
    \end{subfigure}
    \caption{EOD comparison for original embedding (blue) vs. VAE debiased embedding (red) for cancer in 2 years prediction when the (a) male,  (b) female patient's cancer in 2 years label is poisoned by X percentage (X = 0, 25, 50, 75, 100).}
    \label{fig:data_poison2}
\end{figure}

\textbf{Results: } As shown in Figure \ref{fig:data_poison}, as the percentage of male patients' cancer in 1-year labels become corrupted, the EOD for the model using the original embeddings without debiasing increases significantly, reaching close to 1, signifying a dramatic increase in unfairness in its predictions. In contrast, the EOD for the model using the debiased embeddings remains relatively stable, with a maximum EOD of 0.046. The same trend is observed for female data poisoning as well. We observe a similar trend for cancer prediction in 2 years, as shown in Figure \ref{fig:data_poison2}, where the EOD for the model using the original embeddings increases as the percentage of corrupted labels rises, while the EOD for the model with the debiased embeddings remains consistently low.

\section{Discussion}

The results of this study are both promising and significant, demonstrating the potential of adversarial debiasing to enhance fairness in medical AI systems. The ability to remove multiple sensitive demographic information, such as sex and age, from 3D CT embeddings while preserving predictive accuracy for crucial clinical tasks, like lung cancer risk prediction, is a major advancement. This finding is especially important for deploying AI in real-world healthcare, where fairness and equity are paramount. Our framework shows that it is feasible to eliminate demographic biases without compromising the performance of models, which has critical implications for making AI systems more transparent, equitable, and trustworthy in clinical settings.

The encoding of demographic information, including sex and age, within medical imaging embeddings raises concerns about both fairness and security. If AI models learn to encode these attributes, they may inadvertently reinforce biases, leading to discriminatory outcomes for specific patient populations. From a security standpoint, malicious actors can exploit this demographic encoding and may target a specific demographic group to undermine the model's performance. 

The VAE-based debiasing framework enhances both fairness and security by removing demographic information from embeddings, reducing the risk of demographic-based attacks. This ensures that sensitive attributes do not unfairly influence predictions while preserving performance on downstream clinical tasks. By mitigating bias and security vulnerabilities, this approach fosters more transparent and trustworthy AI systems for clinical applications.

Despite the promising results, this work has several limitations. While the debiasing process effectively removed sex and age information, it was not evaluated for other sensitive attributes like race or location. Additionally, the framework was tested only on lung cancer risk prediction from 3D CT images. The effectiveness of debiasing may vary based on image complexity and task-specific requirements, necessitating further investigation into the trade-off between debiasing and downstream performance across different medical imaging domains.


In conclusion, our study demonstrates that VAE adversarial debiasing can effectively remove demographic biases from medical image embeddings without compromising predictive performance. This approach has important implications for creating fairer and more secure healthcare AI systems, helping mitigate both ethical and security risks associated with biased models. Moving forward, future work should explore the debiasing of additional sensitive attributes, further validate this method across diverse datasets, and evaluate its impact on other clinical tasks to ensure the robustness and generalizability of this framework.

\clearpage  

\bibliography{midl-samplebibliography}

\appendix
\counterwithin{figure}{section}
\counterwithin{table}{section}

\section{NLST Dataset} \label{app:data}

\begin{table}[bhtp]
\center
\begin{tabular}{c|c}
\hline
\toprule
\multicolumn{2}{c}{\textbf{NLST}} \\
\midrule
Images/Patients & 64038/12498 \\
\toprule
\multicolumn{2}{c}{\textbf{Train Split}} \\
\midrule
Images/Patients & 52696/10299 \\
\midrule
No. of Males (\% of Images) & 31211 (59.2\%) \\
\midrule
Races (Images,\% of Images) & White: (49040, 0.93\%)\\
& Black or African-American: (1939, 0.04\%)\\
& Asian: (659, 0.01\%)\\
& American Indian or Alaskan Native: (175, 0.0\%)\\
& Native Hawaiian or Other Pacific Islander: (97, 0.0\%)\\
& More than one race: (604, 0.01\%)\\
& Participant refused to answer: (113, 0.0\%)\\
& Other: (69, 0.0\%) \\
\midrule
Age (Mean $\pm$ SD) & 61.6 $\pm$ 5.0 \\
\toprule
\multicolumn{2}{c}{\textbf{Test Split}} \\
\midrule
Images / Patients & 11421/ 2199 \\
\midrule
Males / \% of Images) & 6809/59.6\% \\
\midrule
Races (Images, \% of Images) & White: (10732, 0.94\%)\\
& Black or African-American: (330, 0.03\%)\\
& Asian: (153, 0.01\%)\\
& American Indian or Alaskan Native: (45, 0.0\%)\\
& Native Hawaiian or Other Pacific Islander: (24, 0.0\%)\\
& More than one race: (103, 0.01\%)\\
& Participant refused to answer: (24, 0.0\%)\\
& Other: (10, 0.0\%) \\
\midrule
Age (Mean $\pm$ SD) & 61.8 $\pm$ 5.1 \\
\bottomrule
\end{tabular}
\caption{Detailed information on NLST Dataset.}
\label{tab:data}
\end{table}

\section{Latent Space Optimization} \label{app:latent}

\textbf{Experimental Setup: } The latent space dimension is a critical factor influencing the performance of VAEs. To identify the optimal latent space size, we conducted experiments on the NLST dataset, testing nine configurations: 10, 50, 100, 200, 300, 400, 500, 600, and 700. For each configuration, we trained the VAE on the train split to debias both sex and age and evaluated its performance on the tune split by comparing demographic prediction and lung cancer prediction before and after debiasing. The VAE was trained for 100 epochs with a learning rate of 0.0005 for the encoder and decoder and 0.002 for the adversary, using a batch size of 32.

\begin{figure}
    \centering
    \includegraphics[width=\textwidth]{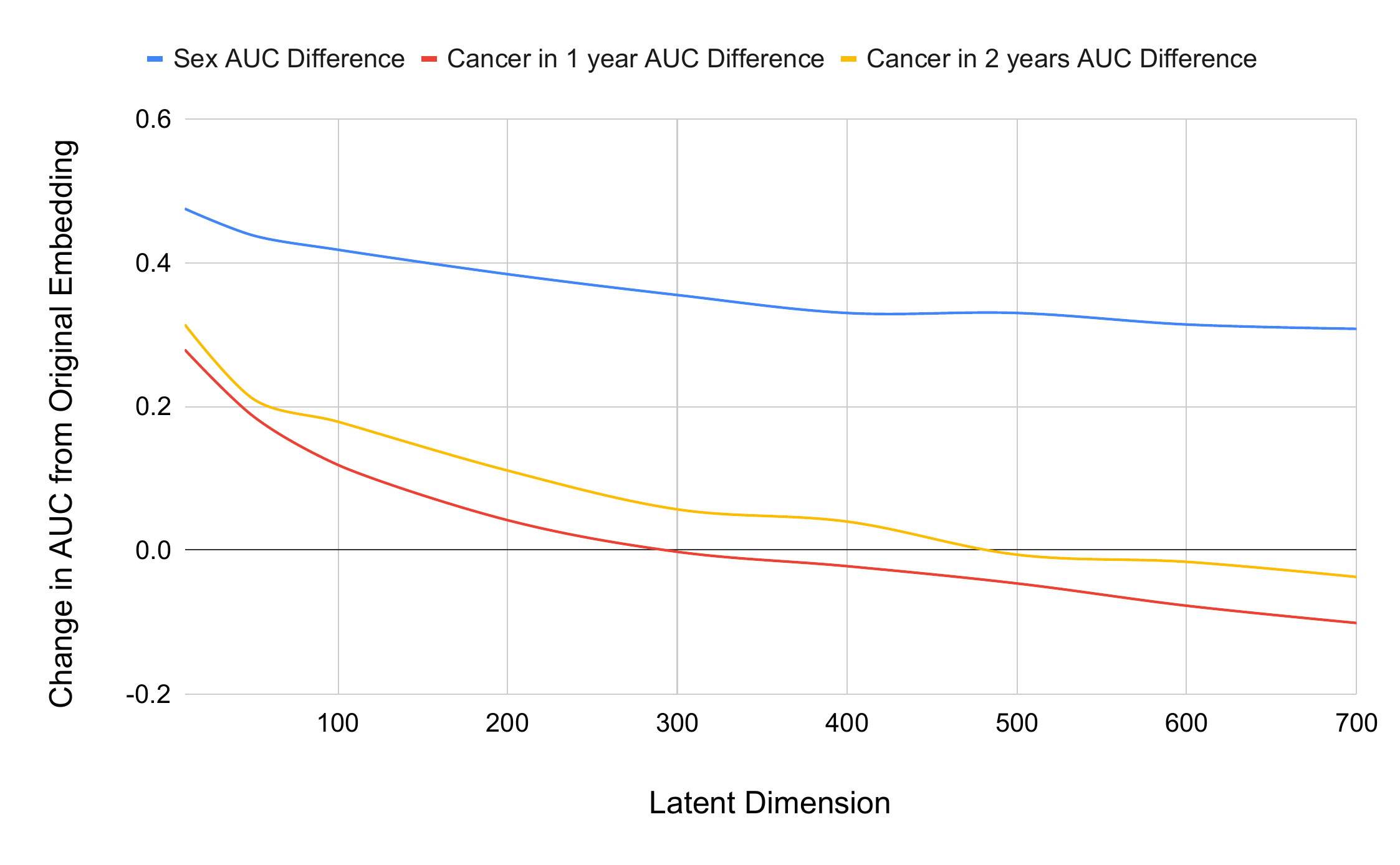}
    \caption{AUC difference between linear classifier prediction of sex (blue), cancer in 1 year (red), and cancer in 2 years (yellow), before - after. The difference for sex is the larger, the better. The difference between cancer in 1 year and cancer in 2 years is the smaller, the better.}
    \label{fig:latent}
\end{figure}

\textbf{Results: } As shown in Table \ref{tab:latentdimension} and Figure \ref{fig:latent}, increasing the latent dimension reduces the AUC difference between the linear classifier predictions for sex, 1-year cancer risk, and 2-year cancer risk before and after debiasing. Larger latent space preserves more information from the original embedding, which benefits downstream task performance but also retains more demographic information. Therefore, identifying an optimal latent dimension is crucial. Based on Table \ref{tab:latentdimension}, we found that at a latent dimension of 500, the reconstructed embeddings maintained or improved performance on lung cancer prediction while significantly reducing performance on sex prediction. Thus, we empirically determined that 500 is the most appropriate choice for balancing predictive performance and demographic debiasing.

\begin{table}[htp]
\center
\resizebox{\textwidth}{!}{%
\begin{tabular}{c|ccccccccc}
\toprule
Latent Dimension & 10 & 50 & 100 & 200 & 300 & 400 & 500 & 600 & 700 \\
\midrule
Sex (AUC) & 0.524 & 0.561 & 0.581 & 0.615 & 0.644 & 0.669 & \textbf{0.669} & 0.685 & 0.691 \\
Age (MAE) & 4.279 & 4.262 & 4.296 & 4.253 & 4.175 & 4.267 & \textbf{4.169} & 4.078 & 4.053 \\
Cancer in 1 Year (AUC) & 0.572 & 0.664 & 0.732 & 0.809 & 0.853 & 0.873 & \textbf{0.897} & 0.928 & 0.952 \\
Cancer in 2 Year (AUC) & 0.536 & 0.639 & 0.671 & 0.739 & 0.793 & 0.81 & \textbf{0.856} & 0.866 & 0.887 \\
\bottomrule
\end{tabular}}
\caption{Performance of linear classifier on debiased embedding produced by VAE of different latent (bottleneck) dimension. Sex AUC lower better, Age MAE higher better, Cancer in 1 year AUC higher better, Cancer in 2 years AUC higher better.}
\label{tab:latentdimension}
\end{table}

\end{document}